\pdfoutput=1

\documentclass[11pt]{article}

\usepackage[preprint]{acl}
\usepackage{bm}
\usepackage{times}
\usepackage{latexsym}
\usepackage{amsmath}
\usepackage{xcolor}
\usepackage{booktabs}
\usepackage{arydshln}
\usepackage[T1]{fontenc}

\usepackage[utf8]{inputenc}

\usepackage{microtype}
\usepackage{booktabs}
\usepackage{arydshln}
\usepackage{inconsolata}

\usepackage{graphicx}

\title{A Lightweight Multi Aspect Controlled Text Generation Solution For Large Language Models}
\author{
    \textbf{Chenyang Zhang~\thanks{Equally Contribution.}}, 
    \textbf{Jiayi Lin~\footnotemark[1]}, 
    \textbf{Haibo Tong}, 
    \textbf{Bingxuan Hou},
    \\
    \textbf{Dongyu Zhang},
    \textbf{Jialin Li},
    \textbf{Junli Wang\thanks{Corresponding author.}} 
    \\
     Tongji University \\
    \texttt{\{inkzhangcy,2331908,2151130,2052643,yidu,2233032,junliwang\}@tongji.edu.cn}\\
}

\begin{document}
\maketitle
\begin{abstract}
Large language models (LLMs) show remarkable abilities with instruction tuning.
However, they fail to achieve ideal tasks when lacking high-quality instruction tuning data on target tasks.
Multi-Aspect Controllable Text Generation (MCTG) is a representative task for this dilemma, where aspect datasets are usually biased and correlated.
Existing work exploits additional model structures and strategies for solutions, limiting adaptability to LLMs.
To activate MCTG ability of LLMs, we propose a lightweight MCTG pipeline based on data augmentation.
We analyze bias and correlations in traditional datasets, and address these concerns with augmented control attributes and sentences.
Augmented datasets are feasible for instruction tuning.
In our experiments, LLMs perform better in MCTG after data augmentation, with a $20\%$ accuracy rise and less aspect correlations.

\end{abstract}
\section{Introduction}
Large language models (LLMs) exhibit ideal abilities in various natural language processing tasks~\citep{lan_are_few_shot_learners,lan_are_zero_shot_reasoners,qin-etal-2023-chatgpt,wei2022emergent,Predictability_and_Surprise_in_Large_Generative_Models}.
LLMs rely on ideal training datasets for task performance enhancement, especially for instruction tuning (IT)~\citep{bai2022traininghelpfulharmlessassistant,touvron2023llamaopenefficientfoundation,JMLR:v25:23-0870} dataset.

However, LLMs struggle on certain downstream tasks since the absence of high-quality IT datasets.
MCTG task suffers from this dilemma.
Existing work~\citep{PPLM,qian-etal-2022-controllable} relies on combinations of single-aspect datasets for supervised learning, which fails to achieve the ideal performance due to issues like aspects bias and correlations~\citep{gu-etal-2022-distributional,liu-etal-2024-multi}.

Recent work addresses corresponding issues through designed models structures ~\citep{carlsson-etal-2022-fine,liu-etal-2024-multi,gu-etal-2022-distributional,yang-etal-2023-tailor}.
Unfortunately, LLMs have enormous model parameters and complex generation process, which is costly to adapt to existing approaches.

In this work, we propose a lightweight MCTG solution for LLMs from the perspective of instruction tuning datasets.
We analyze concerns in existing MCTG datasets and address them in a LLM-based data augmentation pipeline.
For control attributes in existing datasets, different aspect datasets may possess intersection parts.
Attributes are provided in limited label spaces, inaccurate labels fail to recall certain knowledge of models.
For sentences in existing datasets, they exhibit bias which is introduced in dataset construction.
To address aspect bias and correlation, we conduct data augmentation by prompting advanced LLMs.
For control attributes, we provide labels in other aspects for existing sentences and obtain fine-grained accurate attribute descriptions.
For sentences, we rewrite heterogeneous sentences for corresponding control attributes.
We provide mechanisms to ensure effectiveness, diversity and quality of augmentation.
Form of augmented datasets is consistent with original datasets, which can be conveniently transformed into IT datasets.
Consequently, data augmentation is beneficial to common LLMs without specific structures.

We validate the effectiveness of our experiments for LLMs up to 3B scale.
The result shows that the augmented dataset contributes to performance of MCTG, especially exhibiting a more balanced performance for various aspects.
We additionally test mutual information of 3 aspects in generation, result shows that data augmentation diminishes correlations among aspects.
\begin{figure*}[!t]
\centering
\includegraphics[width=0.85\textwidth]{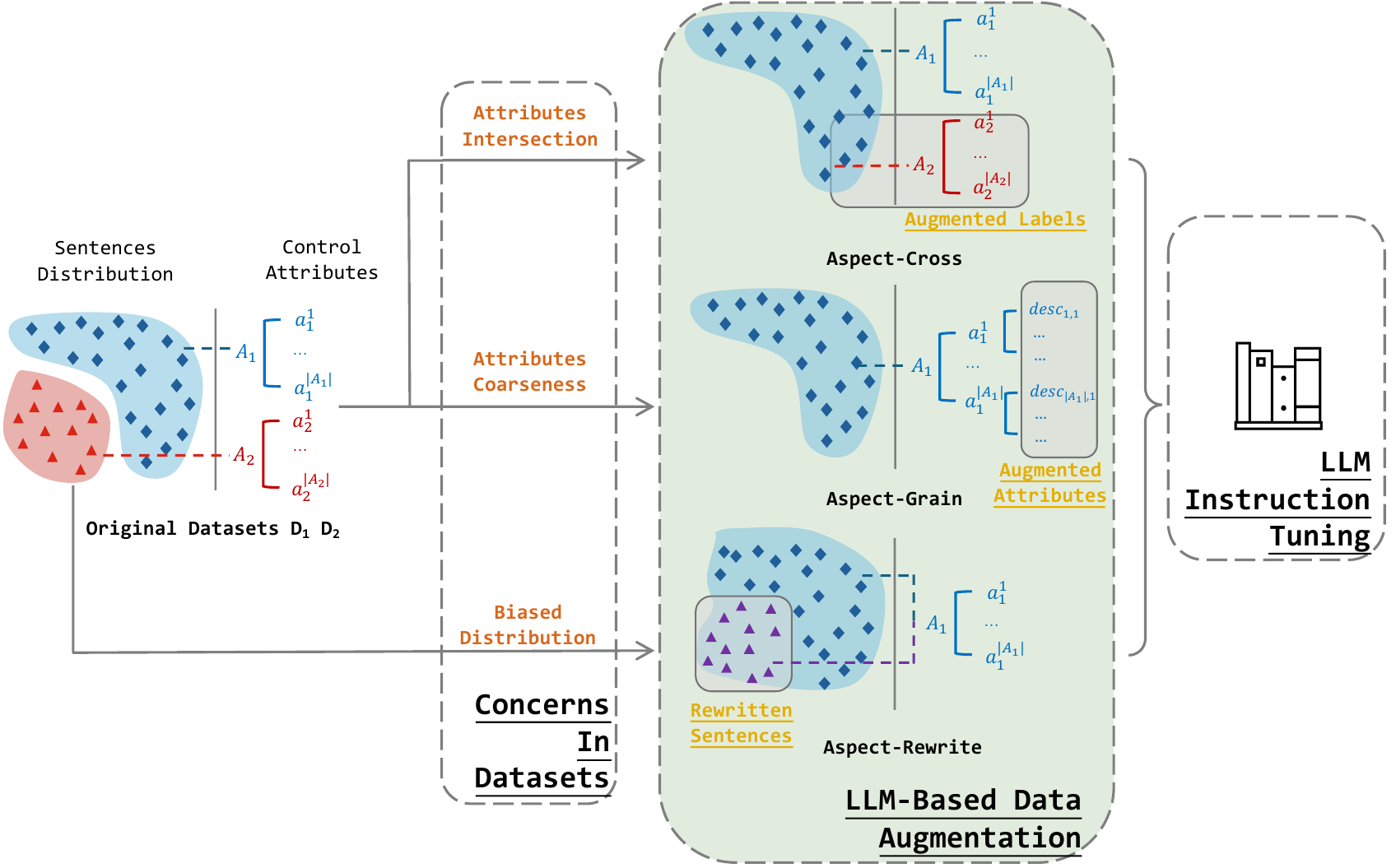}
\caption{An overview of our lightweight MCTG solution.}
\label{fig1:env}
\end{figure*}
\section{Task Formulation}

\paragraph{Control Aspects And Attributes}
For MCTG tasks, controls may contain various $n$ aspects $A=\{A_1,\ldots,A_n\}$.
The $i$-th aspect contains $|A_t|$ exclusive attributes $\{a_{i}^1,\ldots,a_{i}^{|A_t|}\}$\citep{liu-etal-2024-multi}.

MCTG requires a control combination, which selects one attribute from each aspect.
The combination can be notated as a vector of attribute indices $\bm{c} =[c_1,\ldots,c_n]$, where $c_i \in \{1,\ldots,|A_i|\}$ stands for attribute index of $i$-th aspect.

\paragraph{Generation Task Formulation}
With the input of control combinations $\bm{c}$ and generation prompt $m$, generation of language model $LM$ should follow multiple control aspects, notated in Eq.~\ref{eq:sec_task_eq_mtcg}.

\begin{equation}
    LM(m|\bm{c})\sim(a_1^{c_1},\ldots,a_n^{c_n})
    \label{eq:sec_task_eq_mtcg}
\end{equation}

\paragraph{Dataset Application in MCTG}
Existing MCTG tasks are trained on a set of single aspect datasets.
As for $i$-th aspect, training set $\mathcal{D}_i$ is composed of sentences $x$ with its corresponding attribute label $y$ in aspect $A_i$, notated in Eq.~\ref{eq:sec_singleaspect}.
\begin{equation}
    \mathcal{D}_i = \{(x,y) | x\sim (a_i^y), 1 \le y \le |A_i| \}
    \label{eq:sec_singleaspect}
\end{equation}
\section{Methodology}
As shown in Fig.~\ref{fig1:env}, we first analyze 3 representative concerns in existing MCTG datasets.
Then we propose an LLM-based data augmentation pipeline to address the 3 issues correspondingly.
Finally, augmentation data is transformed into format of IT data, for instruction tuning of LLMs.
\subsection{Concerns In Existing MCTG Dataset}
\paragraph{Concerns in Control Attributes}
Attributes from different aspects may share some common concepts, notated as \textbf{attributes intersection}.
For example, IMDB~\cite{IMDB} demonstrates attributes positive and negative in sentiment aspect.
Unfortunately, negative includes toxic attributes like sarcasm for detoxification aspect.

Secondly, control attributes $a_i^t\in A_i$ are predefined, which is not specific and accurate, notated as \textbf{attributes coarseness}.
Taking AGNews~\cite{AGNEWS} as an instance, it provides control aspects of \textit{topic} only in four choices: \textit{Sci/Tech}, \textit{Sports}, \textit{World} and \textit{Business}.
\textit{World} consists of various sub-topics, and sentences inside training set struggle to cover all of the world news, which integrates the bias.
General and ambiguous control attributes obstruct further application on LLMs.
\paragraph{Concerns in Sentences Distributions}
Selections of sentences $x$ in training set are not uniform, with \textbf{biased distribution}.
Distribution of $x$ is biased during dataset construction.
For example, IMDB datasets provide sentences with negative and positive sentiments through crawling movie reviews.
But corresponding control attributes may have instances other than movie reviews, limiting generalization of models.
\subsection{LLM-Based Data Augmentation Pipeline}
We propose a data augmentation pipeline, addressing aforementioned concerns in MCTG datasets~\footnote{In practice, we prompt ~\href{https://platform.openai.com/docs/models/gpt-3-5-turbo}{GPT-3.5-Turbo-0125} for data augmentation, detailed prompts can be found in Appendix.~\ref{sec:augumentation_prompts}.}.
\subsubsection{Aspect-Cross Augmentation}\label{sec:method_aspect_cross}
To address attribute intersection, we exploit LLMs to assign label $\tilde{y}$ in other aspects.
We prompt an advanced LLM for dataset generation.
Augmented dataset is described in Eq.~\ref{eq:sec_aspect_cross_formulation}.
\begin{align}
    \text{cross}(\mathcal{D}_i) &= \{(x,\tilde{y}) | x\sim (a_j^{\tilde{y}}), \nonumber \\
    &\quad 1 \le \tilde{y} \le |A_j|, j\neq i \}
    \label{eq:sec_aspect_cross_formulation}
\end{align}
\paragraph{Contrasting In-Context Learning Demonstrations}
Though LLMs exhibit ability for zero-shot natural language processing, direct prompting is always not trustworthy.
To avoid bias in labeling, we randomly sample examples for every target aspect in each prompt, known as in-context learning (ICL) examples~\cite{lan_are_few_shot_learners}.

\paragraph{Reject Options}
To enhance labeling confidence, we allow LLM to reject~\footnote{For example, LLM can output \textit{None} in response.} for formidable scenarios.
We will neglect all rejected options since some cross aspect labeling is not reasonable.

\paragraph{Consistency Validation}
Considering randomness of LLMs, we repeat each prompt for 3 times and collect all answers.
After normalization of case and format, we only keep consistent responses.
\subsubsection{Aspect-Grained Augmentation}
The development of LLM provides an opportunity to address control coarseness.
We extract unrestricted control attributes for input sentences, extrapolating the label space.
For $\mathcal{D}_i$, we regenerate detailed attribute $desc(x,a_i^y)$ for sentence $x$ with original attribute $a_i^y$. This process is demonstrated in Eq.~\ref{eq:sec_aspect_grained}.
Taking sentiment aspect as an instance, aspect-grained augmentation provides a detailed sentiment like \textit{disappointed} instead of \textit{negative}.
\begin{equation}
    \text{grained}(\mathcal{D}_i) = \{(x,desc(x,a_i^y)) | x\sim desc(x,a_i^y)\}
    \label{eq:sec_aspect_grained}
\end{equation}
In practical prompting, we provide sentences and original control attributes. LLMs are instructed to output detailed descriptions of given attributes but with rejected options.
\subsubsection{Aspect-Rewrite Augmentation}
For concerns in sentence distribution, we rewrite sentences outside current aspect $\tilde{x} \notin \mathcal{D}_i$ with control attribute in $A_i$, as notated in Eq.~\ref{eq:sec_aspect_rewrite}.
The rewritten sentences extrapolate imbalanced distribution in original dataset.
\begin{align}
    \text{rewrite}(\mathcal{D}_i) &= \{(\tilde{x}, y) \,|\, \tilde{x} \sim (a_i^y),\nonumber \\
    &\quad 1 \le y \le |A_i|, \, \tilde{x} \notin \mathcal{D}_i \}
    \label{eq:sec_aspect_rewrite}
\end{align}
In practice, we select sentences in other aspects and rewrite them with current aspect controls, with contrastive ICL examples and rejected options.

\paragraph{Quality control} We eliminate instances that evidently deviate from statistical norms (i.e. very short sentences).
Additionally, we filter unsuccessful rewriting due to the task difficulty.
In practice, LLMs may copy the input or output abnormal responses.
We compare semantic similarity~\footnote{We use \href{https://huggingface.co/BAAI/bge-large-en-v1.5}{bge-large-en-v1.5} as semantic embedder and calculate the cosine similarity between two sentences.} before and after rewriting, then eliminate top $50\%$ and bottom $10\%$ of similar instances.
\subsection{Instruction Tuning Dataset Construction}\label{sec:SFT_CONS}
Augmented datasets share common format with original datasets, and we transform them into IT dataset for training.
An instance of IT dataset consist of instruction $I$ and response $R$.
LLMs should output $R$ with the input of $I$.

For an instance $(x,y) \in \mathcal{D}_i$, we provide simple task descriptions, target control attribute $a_i^y$, and generation prefix~\footnote{In~\citet{gu-etal-2022-distributional,PPLM}, MCTG should start with certain generation prompts. Consequently, downstream evaluation experiments contain corresponding prefix. We also provide this requirement in instruction, like \textit{starting with $p(x)$} (standing for first three words of response).} in $I$.
We simply use controlled sentence $x$ as $R$. An instance is in Appendix.~\ref{apx:ITdetails}.
\section{Experiments}
\begin{table*}
\small
\centering
\begin{tabular}{ccccc} 
\toprule
\textbf{Baselines} & \textbf{Total Accuracy↑(\%)} & \textbf{Sentiment↑(\%)} & \textbf{Topic↑(\%)} & \textbf{Detoxification↑(\%)}  \\ 
\hline
Augmented MCTG     & \textbf{47.57}               & 77.75                   & 71.11               & 82.75                         \\
w/o Cross          & 44.03                        & 77.32                   & 61.46               & 85.39                         \\
w/o Grain          & 35.25                        & 84.36                   & 59.89               & 71.18                         \\
w/o Rewrite        & 29.67                        & 93.27                   & 55.61               & 59.68                         \\ 
\hdashline
Vanilla MCTG       & 22.14                        & 98.86                   & 41.89               & 51.35                         \\
\bottomrule
\end{tabular}
\caption{Overall result on MCTG, best total accuracy is bold. Accuracy indicates ratio of controlled sentences evaluated by classifiers in ~\citet{gu-etal-2022-distributional}. \textbf{Total accuracy} indicates ratio of generations fit all 3 control aspects.}
\label{tab:CTGResult}
\end{table*}
\subsection{Datasets Selection}
\paragraph{Basic Datasets}
Following~\citet{gu-etal-2022-distributional}, we select IMDB~\cite{IMDB}, AGNews~\cite{AGNEWS} and Jigsaw Toxic Comment~\footnote{\href{https://www.kaggle.com/c/jigsaw-toxic-comment-classification-challenge/}{https://www.kaggle.com/c/jigsaw-toxic-comment-classification-challenge/}.} for sentiment, topic and detoxification aspects.
\paragraph{Augmented Datasets}
We conduct aspect-cross augmentation for each two of basic datasets, and aspect-grained augmentation for all of basic datasets.
For aspect-rewrite augmentation, we select each aspect and rewrite sentences of the other two aspects for current aspect control~\footnote{Detoxification is skipped in rewriting, since GPT-3.5 is aligned not to generate harmful expressions.}.
\paragraph{Datasets Mixture}
Distributions of IT datasets influence LLM performance~\cite{instag}.
We integrate universal IT datasets with MCTG datasets, to avoid overfitting and instruction-ability degradation.
For baselines, \textbf{Augmented MCTG} consists of universal IT data, vanilla MCTG dataset, and all augmented MCTG data. While \textbf{Vanilla MCTG} replaces augmented data with incremental universal IT datasets.
Details are in Appendix.~\ref{apx:datastat}.

\subsection{Model Training}
We apply Qwen-2.5-3B~\cite{qwen2}~\footnote{\href{https://huggingface.co/Qwen/Qwen2.5-3B}{https://huggingface.co/Qwen/Qwen2.5-3B}} as LLM backbone, and conduct LoRA~\cite{lora} during tuning.
Details are in Appendix.~\ref{apx:training_settings}.

\subsection{Evaluation}
We experiment with the same control combinations, prefix and evaluation models to~\citet{gu-etal-2022-distributional,pascual-etal-2021-plug-play}.
We additionally repeat each generation $10$ times and set temperature to $0.2$ for LLMs to weaken randomness.
\subsection{Experiment Results}
\paragraph{MCTG Performance}
As shown in Table.~\ref{tab:CTGResult}, augmented MCTG datasets enhance the performance of MCTG, especially in total combinations and certain aspects.
Augmented MCTG datasets enhance the total accuracy significantly($20\%$).
Original datasets have a bias on sentiment aspects, and neglect the learning of the other two aspects due to unprocessed aspect correlations and bias.
Augmented datasets successfully address these concerns and re-balance three aspects in the generation.
Therefore, total and each aspect accuracy are enhanced.
As for ablation study, aspect rewrite is the most influential one for performance, which indicates LLMs are more sensitive to sentence features during instruction tuning.
Sentences are trained as response so that more uniform and diverse responses are beneficial for LLMs in MCTG.
In Appendix.~\ref{apx:case_study}, we conduct a case study on for model generations.
\paragraph{Aspect Correlations}
To demonstrate aspect correlations learned by LLMs, we record predicted attributes distribution and their mutual information (MI)~\cite{MIsha,mi2}.
We calculate MI of all three aspects and each two of them, results are shown in Table.~\ref{tab:MI_Result}.
Control attributes are combined orthogonally in instructions, so ideal MI items should be $0$.
Augmented MCTG weakens correlations among aspects, but two baselines still share an identical trend of correlations, necessitating further processing with correlations.
\begin{table}
\small
\centering
\refstepcounter{table}
\label{LanguageModeling}
\begin{tabular}{ccc} 
\toprule
\ & Augmented MCTG & Vanilla MCTG  \\ 
\hline
MI($A_1$,$A_2$,$A_3$) & 0.280          & 0.508        \\
MI($A_1$,$A_2$)   & 0.042          & 0.173        \\
MI($A_1$,$A_3$)   & 0.231          & 0.331        \\
MI($A_2$,$A_3$)   & 0.016          & 0.074        \\
\bottomrule
\end{tabular}
\label{tab:MI_Result}
\caption{MI of three aspects for generations. $A_1$,$A_2$,$A_3$ stand for sentiment, topic and detoxification aspects.}
\end{table}
\section{Conclusion}
In this work, we construct a lightweight MCTG solution for LLMs.
Starting from a perspective of datasets, we analyze concerns in traditional MCTG datasets including attributes intersection, attributes coarseness, and biased distribution.
Then we provide a LLM-based data augmentation pipeline for better instruction tuning datasets, including generating cross labels, generating fine-grained label descriptions and rewriting heterogeneous sentences for target aspects.
In experiments, training LLM with augmented data exhibits enhanced and balanced performances on overall aspects.
The result indicates our solution is effective for LLMs.
Result of MI shows that augmented dataset weakens correlations between certain aspects.

\section{Limitations}
In this work, we propose a lightweight solution to activate MCTG ability for LLMs.
Our work still leaves some limitations for future discussion as follows:

(1) The data augmentation pipeline relies on advanced LLMs like GPT3.5, which is a compromising option for complex data synthetic tasks~\cite{personahub,yang-etal-2023-refgpt}. But a self-conditioned augmentation pipeline is more feasible for lightweight solutions, where data augmenter LLMs and trained LLMs remain the same like self-distill~\cite{dubey2024llama,xu-etal-2023-baize}.

(2) The quality control of augmentation relies on a strict and simple filter policy, we expect for more explainable filter strategies to enhance data productivity.

(3) Our work focuses on instruction tuning of LLMs for MCTG, but leaves other post-training processes like RLHF~\cite{ouyang2022traininglanguagemodelsfollow} and DPO~\cite{DPO} for future discussions.
\section{Ethical Considerations}
In this work, the trained MCTG model includes a toxic aspect, which may result in the generation of toxic content during evaluation. However, the inclusion of the toxic aspect is solely for the purpose of evaluating the model's capabilities. We assure that we will not require the model to generate toxic content in real-world applications.

\bibliography{custom}

\appendix
\definecolor{c1red}{RGB}{192,0,0}
\definecolor{c2blue}{RGB}{0,112,192}
\definecolor{c3green}{RGB}{0,176,80}
\definecolor{c4purple}{RGB}{112,48,160}
\definecolor{c5pink}{RGB}{255,102,153}

\section{Related Work}

\paragraph{Large Language Models}
Large language models (LLMs), such as LLaMA~\citep{touvron2023llamaopenefficientfoundation,dubey2024llama} and GPT-4~\citep{achiam2023gpt}, refer to a series of Transformer-based models undergoing extensive pretraining with massive corpora. By scaling up the data volume and model capacity, LLMs demonstrate remarkable emergent capabilities, such as In-Context Learning (ICL)~\citep{brown2020language} and Chain-of-Thought (CoT) prompting~\citep{wei2022chain}, enable them to comprehend human instructions and handle  complex tasks with minimal or even no supervision. Despite their exceptional performance, LLMs still produce nonsensical or incongruent information in practical applications (e.g. "hallucination"\citep{ji2023survey}). In this paper, our method leverages the knowledge and generative capabilities of LLMs.
\begin{table*}[!h]
\centering
\begin{tabular}{cc} 
\toprule
Baselines                                 & Datasets                                                                                                           \\ 
\hline
Augmented MCTG                     & 28.5k \textbf{Univ.} + 9k \textbf{Vanilla} + 3k \textbf{Cross.~}+ 3k \textbf{Grained.} + 1.5k \textbf{Rewrite.}~  \\
\multicolumn{1}{r}{w/o\textit{ Cross.}}   & 31.5k \textbf{Univ.} + 9k \textbf{Vanilla~}+ 3k \textbf{Grained.} + 1.5k \textbf{Rewrite.}                        \\
\multicolumn{1}{r}{w/o \textit{Grained.}} & 31.5k \textbf{Univ.} + 9k \textbf{Vanilla~}+ 3k \textbf{Cross.} + 1.5k \textbf{Rewrite.}                          \\
\multicolumn{1}{r}{w/o \textit{Rewrite.}} & 30k \textbf{Univ.} + 9k \textbf{Vanilla~}+ 3k \textbf{Cross.} + 3k \textbf{Grained.}                               \\ 
\hline
Vanilla MCTG                     & 36k \textbf{Univ.} + 19k \textbf{Vanilla}                                                                        \\ 

\bottomrule
\end{tabular}
\caption{Training dataset statistics of all baselines in experiments.}
\label{tab:datastat}
\end{table*}
\paragraph{Multi Aspect Controlled Text Generation}
From the perspective of parameter fusion, \citet{huang-etal-2023-extensible} have improved MACTG in prefix tuning\citep{li2021prefix} by adjusting the positions where prefixes are added, thereby reducing the mutual influence of multiple prefixes. Tailor~\citep{yang-etal-2023-tailor} adjust the multi-attribute prompt mask and re-index the position sequence to bridge the gap between the training phase (where each task uses a single-attribute prompt) and the testing phase (where two prompts are connected).

On the other hand, \citet{gu-etal-2022-distributional} approaches this issue from the perspective of distribution within semantic space. After obtaining the intersection of attribute distributions, the language model's distribution is biased toward this region. However, the intersection of different attribute distributions may not overlap. To address this, MacLaSa~\citep{ding-etal-2023-maclasa} estimates a compact latent space to improve control ability and text quality, mitigating interference between different aspects. \citet{liu-etal-2024-multi} propose MAGIC, which uses counterfactual feature vectors in the latent space to disentangle attributes, alleviating the imbalance in attribute correlation during training.

Regarding the scarcity of training data for MCTG, \citet{zhang-etal-2023-macsum} propose MACSUM, a human-annotated dataset containing summaries with mixed control attributes. \citet{chen-etal-2023-mixture} use a strategy of mixing soft prompts to help large models generate training data that aligns with multi-aspect control attributes.

\section{Data Augmentation Prompts}\label{sec:augumentation_prompts}
\begin{figure*}[!h]
\centering
\includegraphics[width=1.0\textwidth]{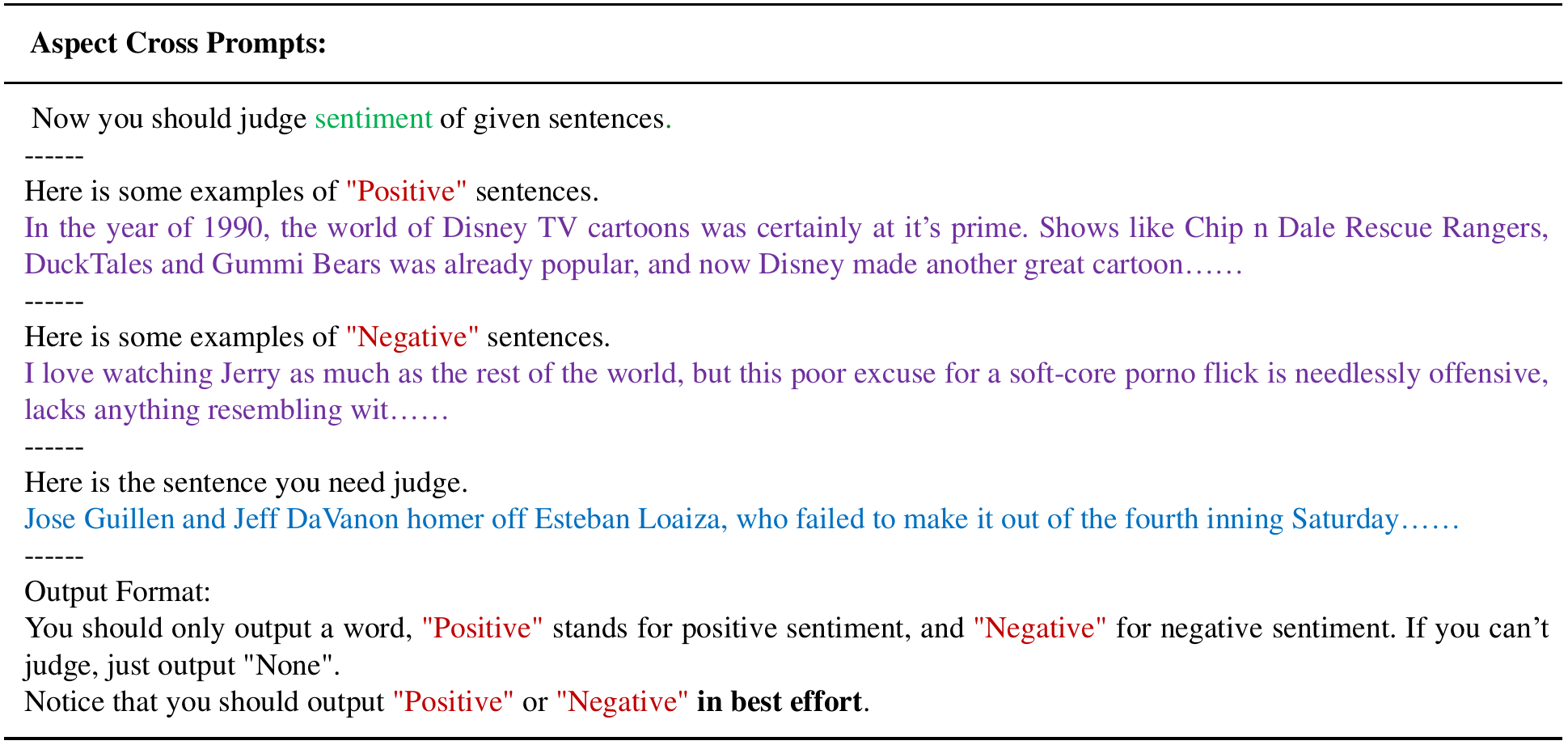}
\caption{The prompt of Aspect-Cross Augmentation}
\label{fig:apx_aspect_cross}
\end{figure*}

\paragraph{Aspect-Cross Augmentation}
Fig.~\ref{fig:apx_aspect_cross} shows the  prompt of Aspect-Cross Augmentation.
\textcolor{c3green}{Aspects descriptions} are colored green; \textcolor{c1red}{attributes descriptions} are colored red;
\textcolor{c4purple}{ICL examples of target attributes} are colored purple; \textcolor{c2blue}{target sentences for label} are colored blue.
Bold fonts are written in markdown format like \textit{**Example**}. 

\begin{figure*}[!h]
\centering
\includegraphics[width=1.0\textwidth]{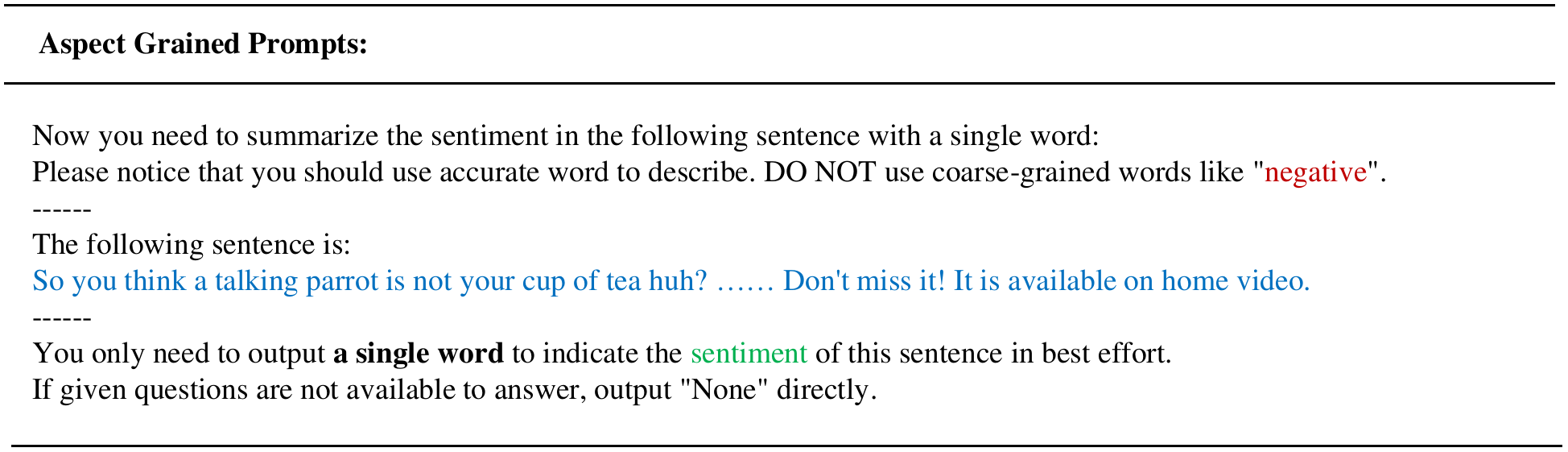}
\caption{The prompt of Aspect-Grained Augmentation}
\label{fig:apx_aspect_grain}
\end{figure*}

\begin{figure*}[!h]
\centering
\includegraphics[width=1.0\textwidth]{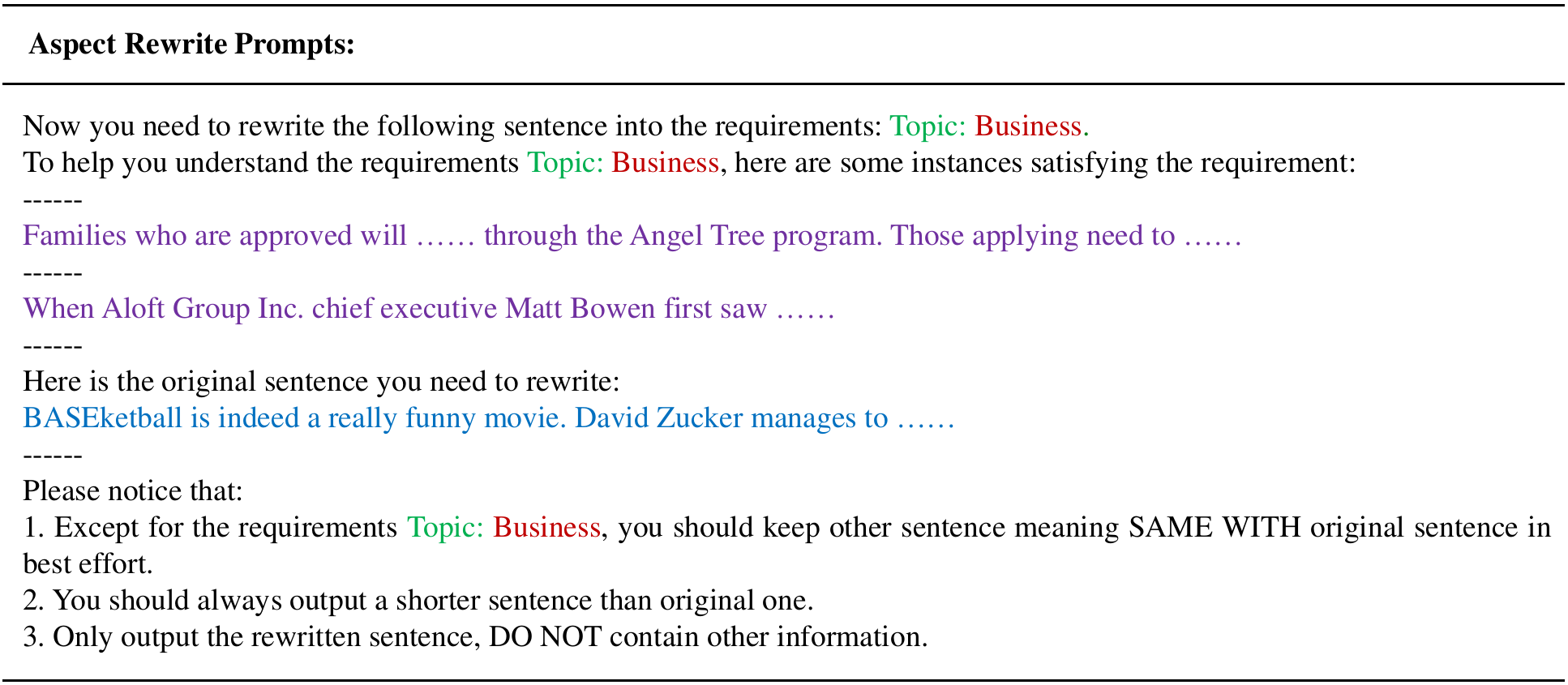}
\caption{The prompt of Aspect-Rewrite Augmentation}
\label{fig:apx_aspect_rewrite}
\end{figure*}

\paragraph{Aspect-Grained Augmentation}
Fig.~\ref{fig:apx_aspect_grain} shows the  prompt of Aspect-Grained Augmentation.
\textcolor{c3green}{Aspects descriptions} are colored green; \textcolor{c1red}{attributes descriptions} are colored red;
\textcolor{c2blue}{target sentences for grained augmentation} are colored blue.
\paragraph{Aspect-Rewrite Augmentation}
Fig.~\ref{fig:apx_aspect_rewrite} shows the prompt of Aspect-Rewrite Augmentation.
\textcolor{c3green}{Aspects descriptions} are colored green; \textcolor{c1red}{attributes descriptions} are colored red;
\textcolor{c4purple}{ICL examples for rewriting} are colored purple;
\textcolor{c2blue}{sentences need to be rewritten} are colored blue.

\section{Details Of Instruction Tuning Dataset Construction}\label{apx:ITdetails}
\begin{figure*}[htbp]
\centering
\includegraphics[width=1.0\textwidth]{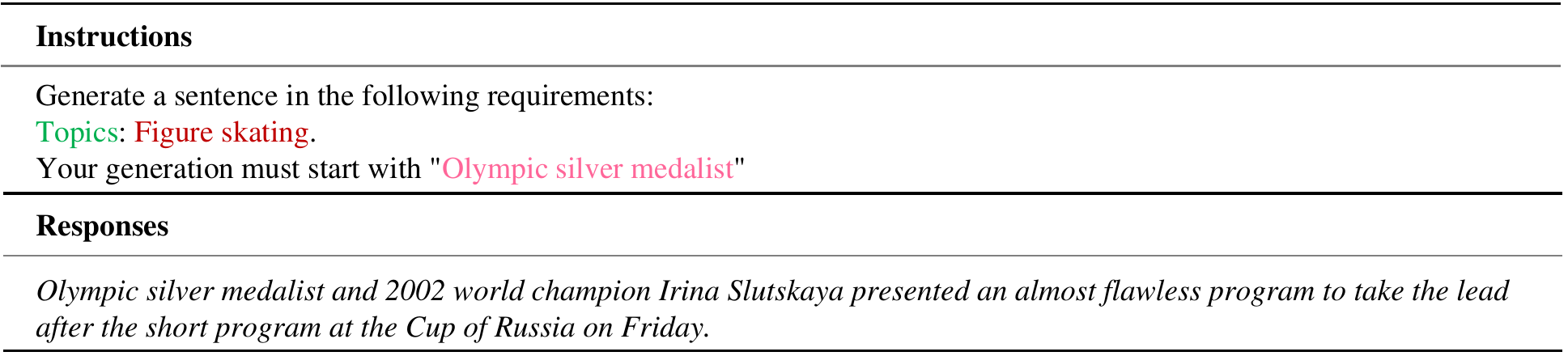}
\caption{An instance of instruction datasets for MCTG.}
\label{fig:IT_dataset_Example}
\end{figure*}

Fig.~\ref{fig:IT_dataset_Example} shows the final instruction and response pair of an IT dataset instance.
\textcolor{c3green}{Aspects descriptions} are colored green; \textcolor{c1red}{attributes descriptions} are colored red; \textcolor{c5pink}{prefixes for generation} are colored pink.

\section{Datasets Statistics}\label{apx:datastat}
In our instruction tuning process, we conduct three categories of datasets as followed:

\paragraph{Augmented Datasets}
Augmented datasets including aspect-cross augmentation (notated as \textbf{Cross.}), aspect-grained augmentation (notated as \textbf{Grained.}) and aspect-rewrite augmentation (notated as \textbf{Rewrite.}).

\paragraph{Universal Instruction Tuning Datasets} (notated as \textbf{Univ.})
We exploit a mixture of Deita-10k-v0~\footnote{\href{https://huggingface.co/datasets/hkust-nlp/deita-10k-v0}{https://huggingface.co/datasets/hkust-nlp/deita-10k-v0}}~\cite{deita}, Airobos3.2~\footnote{\href{https://huggingface.co/datasets/HuggingFaceH4/airoboros-3.2}{https://huggingface.co/datasets/HuggingFaceH4/airoboros-3.2}}, Capybara~\footnote{\href{https://huggingface.co/datasets/LDJnr/Capybara}{https://huggingface.co/datasets/LDJnr/Capybara}}, no-robots~\cite{no_robots} ~\footnote{\href{https://huggingface.co/datasets/HuggingFaceH4/no\_robots}{https://huggingface.co/datasets/HuggingFaceH4/no\_robots}} for universal IT datasets.
They are all popular instruction-tuning datasets in community, whose instructions cover a wide range of universal tasks for LLMs.

\paragraph{Vanilla CTG Datasets} (notated as \textbf{Vanilla})
We exploit original version of IMDB~\cite{IMDB}, AGNews~\cite{AGNEWS} and Jigsaw Toxic Comment, transforming them into IT format like Sec.~\ref{sec:SFT_CONS}.

We conduct random sample on these datasets, to keep dataset volume of each baseline identical.
Final statistics of all baselines are demonstrated in Table.~\ref{tab:datastat}.
\section{Hyperparameter Settings}\label{apx:training_settings}
Hyperparameter settings for instruction tuning and generation are shown in Table.~\ref{tb:hyperparameter}. Training loss is only calculated for response tokens.
We train models on 3 NVIDIA V100 GPUs for 6 hours in each experiment.
\begin{table}
\centering
\small
\begin{tabular}{cc} 
\hline
\textbf{Hyperparameter}       & \textbf{Value}  \\ 
\hline
Learning Rate                  & 5e-5            \\
Learning Rate Scheduler        & Cosine          \\
Warmup Steps                   & 20              \\
Training Batch Size             & 144              \\
Max Input Length               & 3072            \\
Max Generated Length           & 128             \\
Precision of Tensor            & Float32         \\
Vocabulary Size                & 151642          \\
Random Seed                    & 1996            \\
Epochs                         & 2               \\
Optimizer                      & Adam            \\
LoRA Rank                      & 32              \\
LoRA $\alpha$ & 32              \\
LoRA Dropout                   & 0.1             \\
Rank-Stabilized LoRA~\cite{rslora}                        & Enabled   \\
Chat Template                  & ChatML          \\
\hline
\end{tabular}
\caption{Hyperparameter Settings}
\label{tb:hyperparameter}
\end{table}

\section{Case Study}\label{apx:case_study}
\begin{figure*}
\centering
\includegraphics[width=1.0\textwidth]{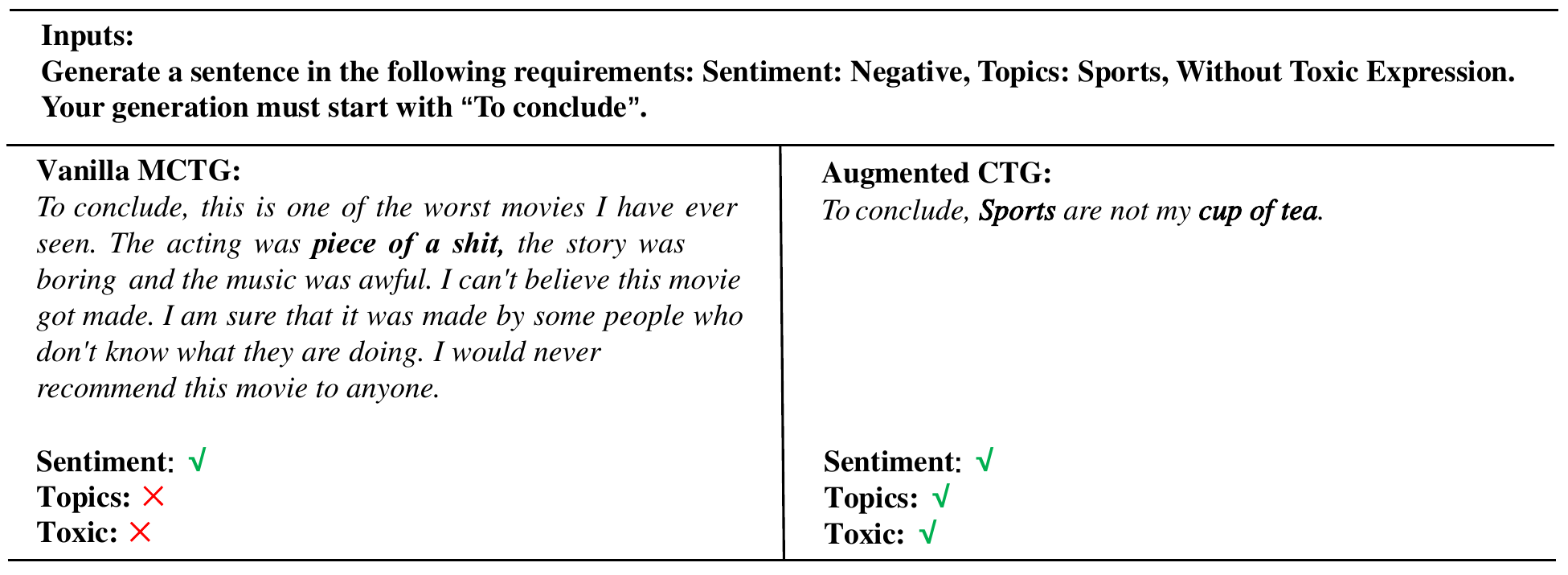}
\caption{A simple case study. Key sentence components demonstrating control attributes are in \textbf{bold}.}
\label{fig:apx_Case_Study}
\end{figure*}

\textcolor{c1red}{\textbf{Warning: This sections may contains offensive and toxic sentences}}. Fig.~\ref{fig:apx_Case_Study} presents a detailed example, where model is required to generating text with a negative sentiment, a title of sports and without toxic expressions. The sentences generated by Vanilla MCTG meet the sentiment requirement but fail to align with the topic and toxic criteria, and these sentences are relatively verbose. In contrast, the sentences generated by Augmented CTG meet all requirements and are more concise and elegant. This indicates that the Augmented CTG method enables the model to generate sentences that better adhere to multiple aspects.

\end{document}